\documentclass[10pt,twocolumn,letterpaper]{article}

\usepackage{iccv}
\usepackage{times}
\usepackage{epsfig}
\usepackage{graphicx}
\usepackage{amsmath}
\usepackage{amssymb}
\usepackage{times}
\usepackage{epsfig}
\usepackage{graphicx}
\usepackage{amsmath}
\usepackage{amssymb}
\usepackage{subcaption}
\usepackage{epsfig}
\usepackage{booktabs}
\usepackage{multirow}

\usepackage[breaklinks=true,bookmarks=false]{hyperref}

\iccvfinalcopy 

\def\iccvPaperID{****} 
\def\httilde{\mbox{\tt\raisebox{-.5ex}{\symbol{126}}}}

\ificcvfinal\fi

\newcommand\blfootnote[1]{%
  \begingroup
  \renewcommand\thefootnote{}\footnote{#1}%
  \addtocounter{footnote}{-1}%
  \endgroup
}

\begin{document}

\def\httilde{\mbox{\tt\raisebox{-.5ex}{\symbol{126}}}}


\title{Fast and Efficient Model for Real-Time Tiger Detection In The Wild}

\author{Orest Kupyn \thanks{These two authors contributed equally} \\
Ukrainian Catholic University\\
Lviv, Ukraine\\
{\tt\small kupyn@ucu.edu.ua}
\and
Dmitry Pranchuk\footnotemark[1]
\\
WANNABY\\
Minsk, Belarus\\
{\tt\small d.pranchuk@gmail.com}\\
}
\twocolumn[{%
\renewcommand\twocolumn[1][]{#1}%
\maketitle
\begin{center}
    \centering
    \setlength{\tabcolsep}{0.3mm}
    \renewcommand{\arraystretch}{0.25}
    \vspace{-8mm}
    \begin{tabular}{cccc}
    \includegraphics[width=4.3cm]{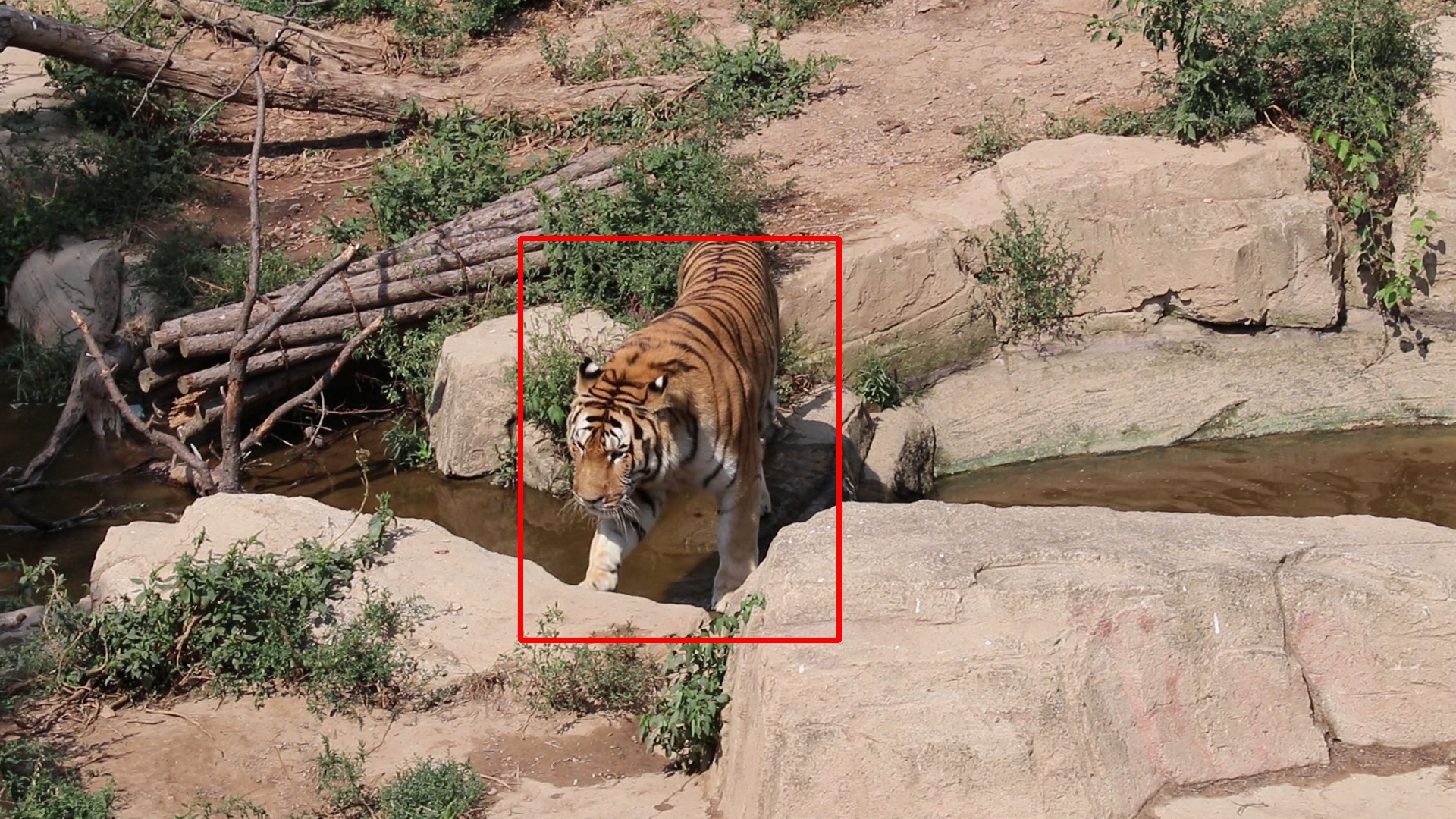} &
    \includegraphics[width=4.3cm]{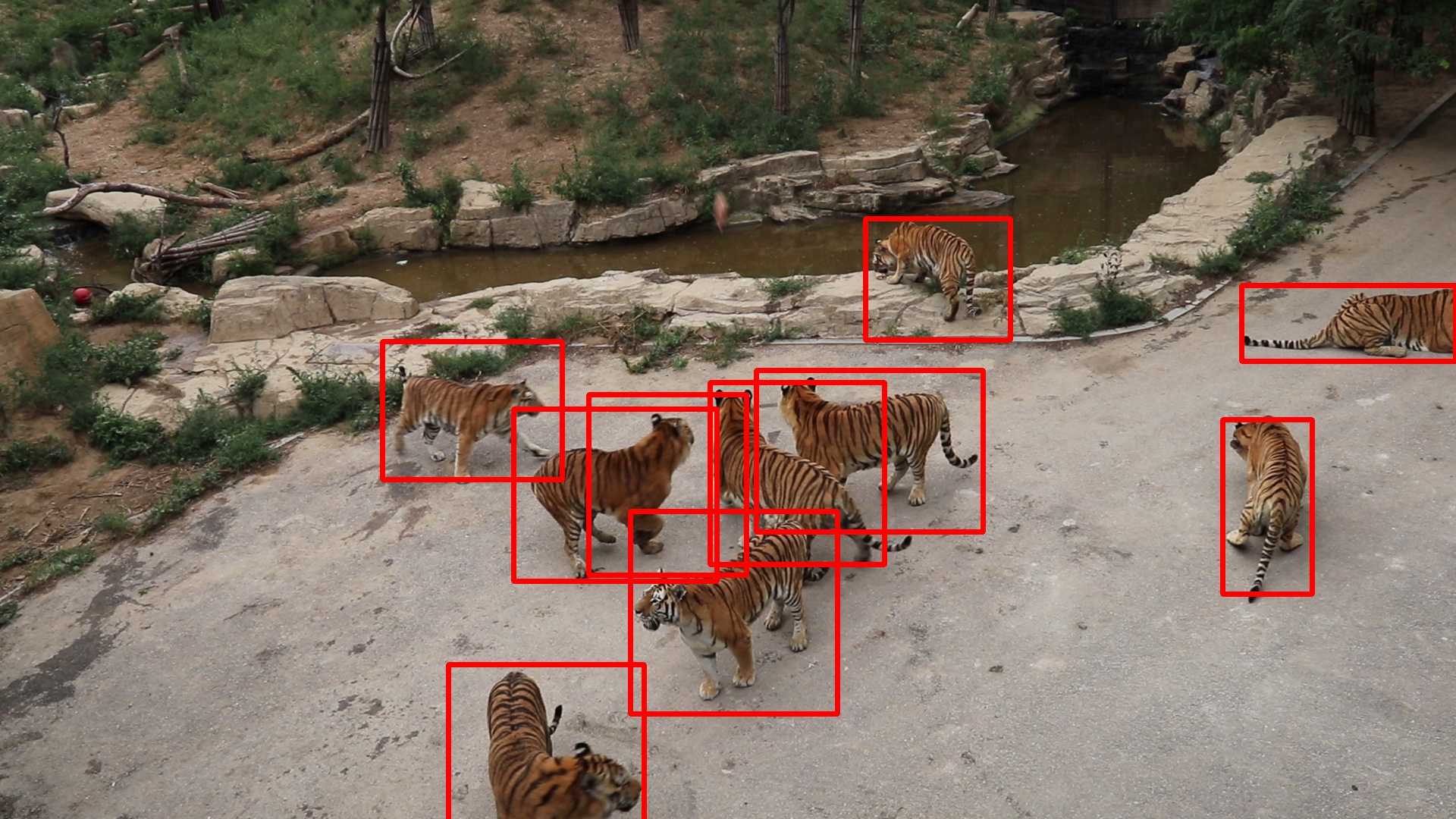} &
    \includegraphics[width=4.3cm]{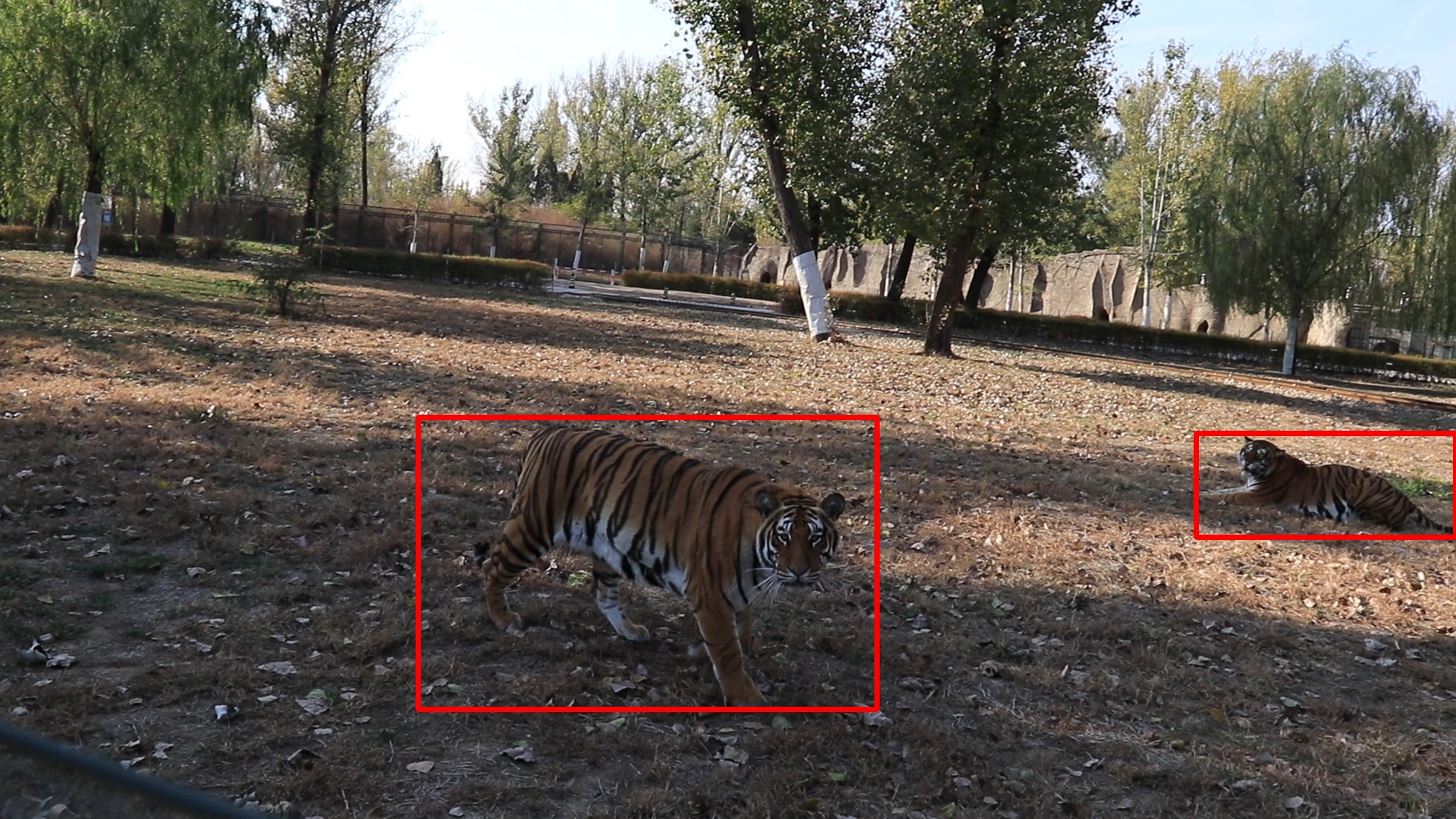} &
    \includegraphics[width=4.3cm]{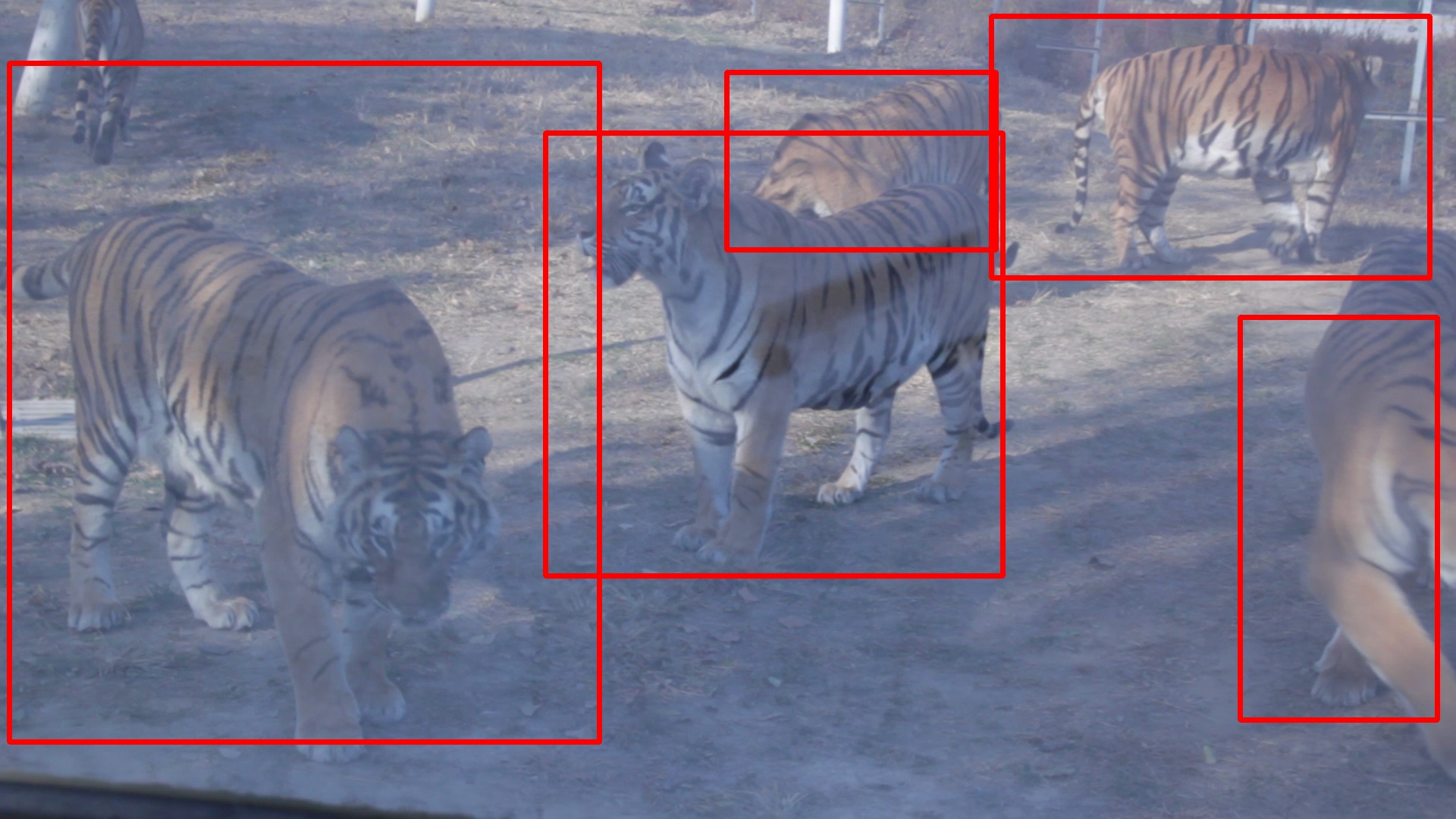}
    \end{tabular}
    \begin{tabular}{cccc}
    \includegraphics[width=4.3cm]{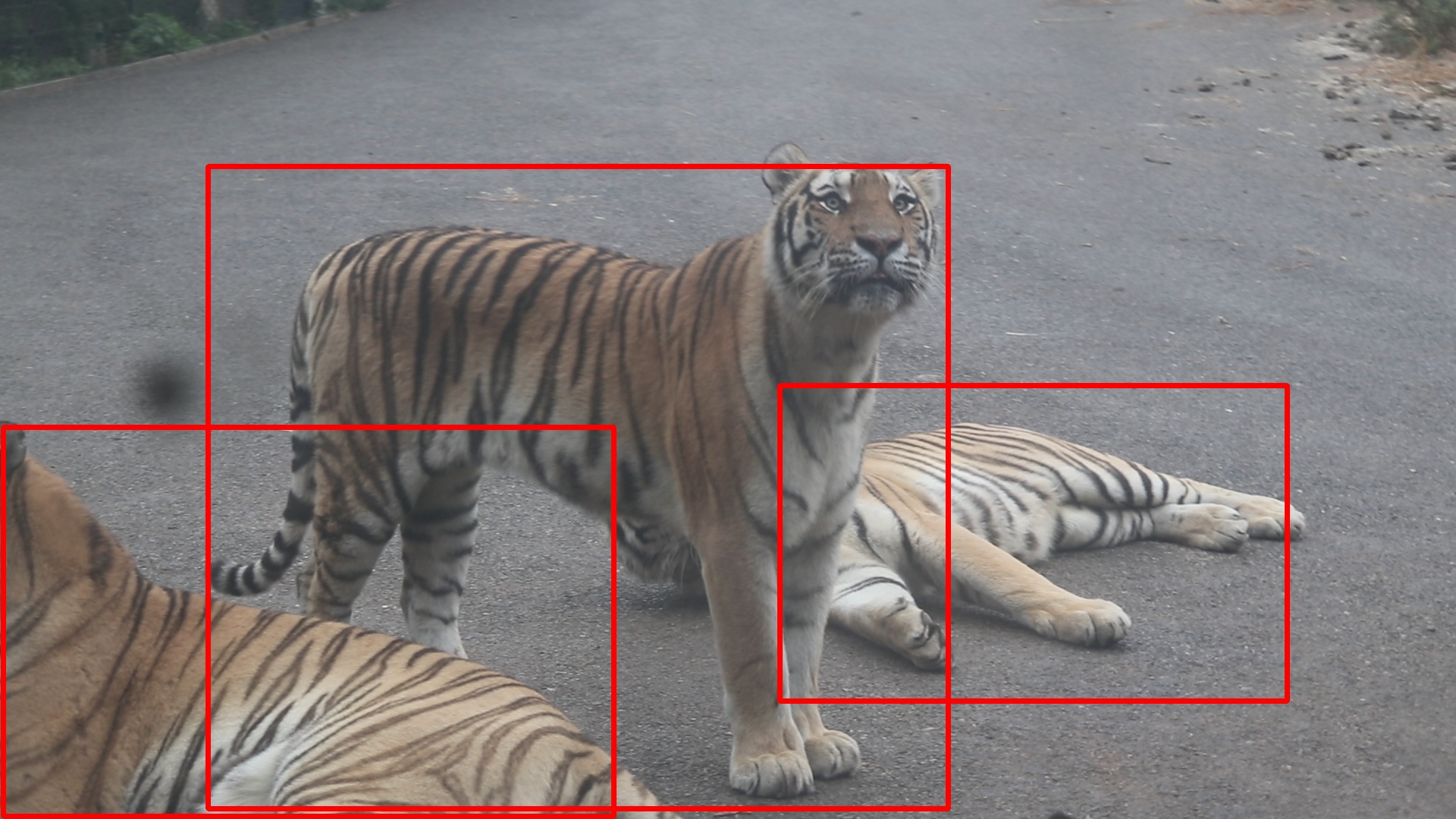} &
    \includegraphics[width=4.3cm]{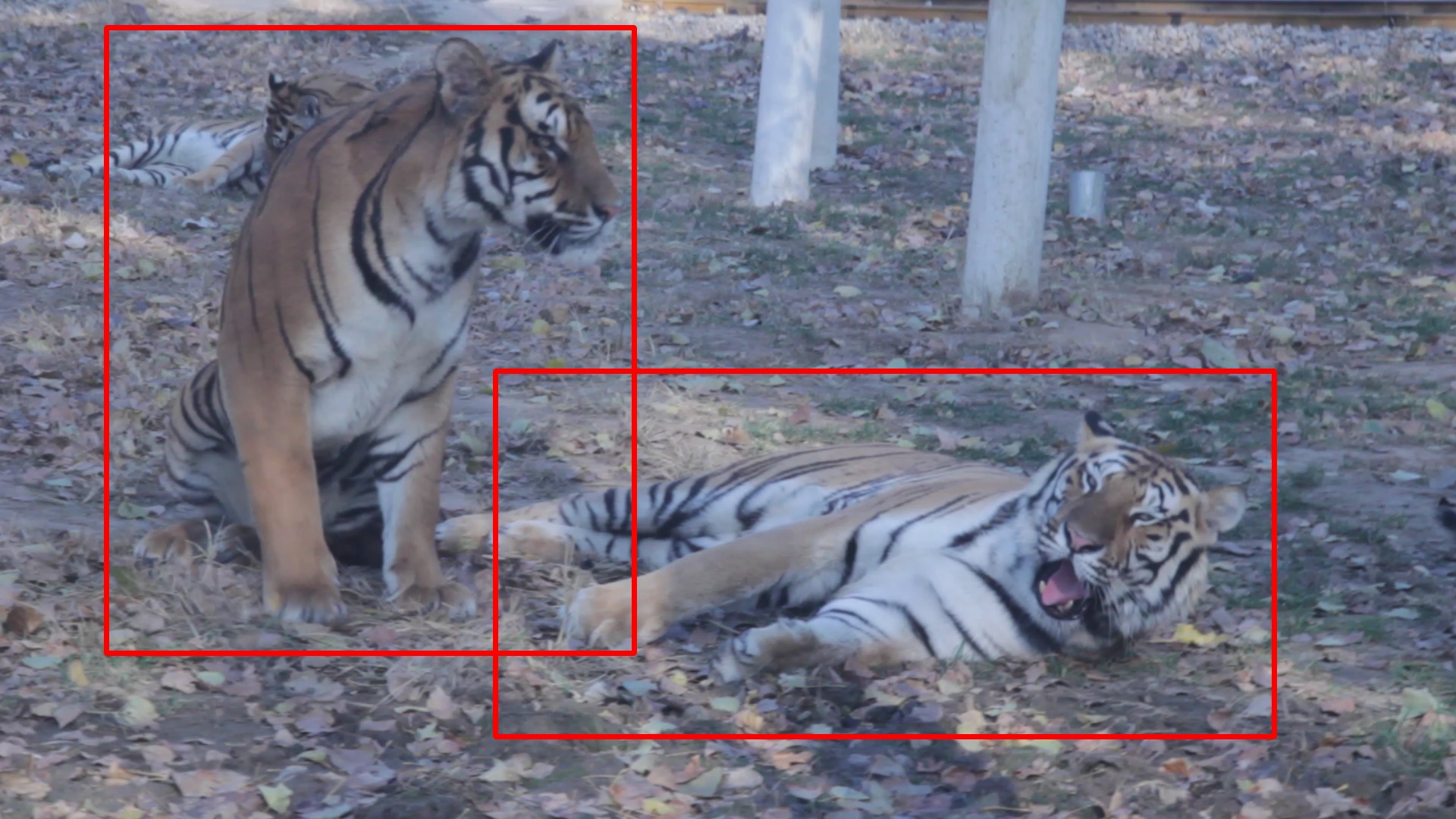} &
    \includegraphics[width=4.3cm]{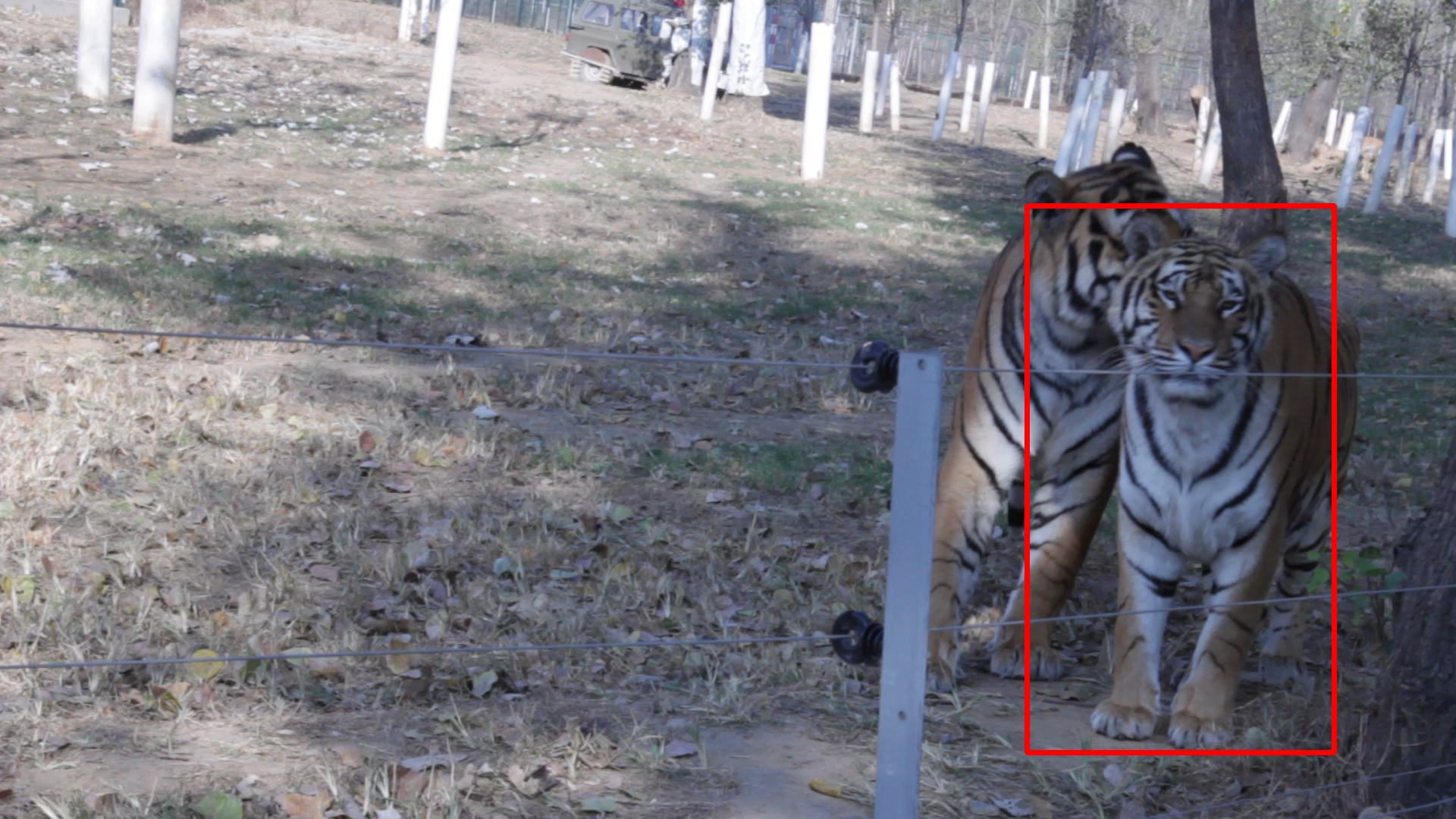} &
    \includegraphics[width=4.3cm]{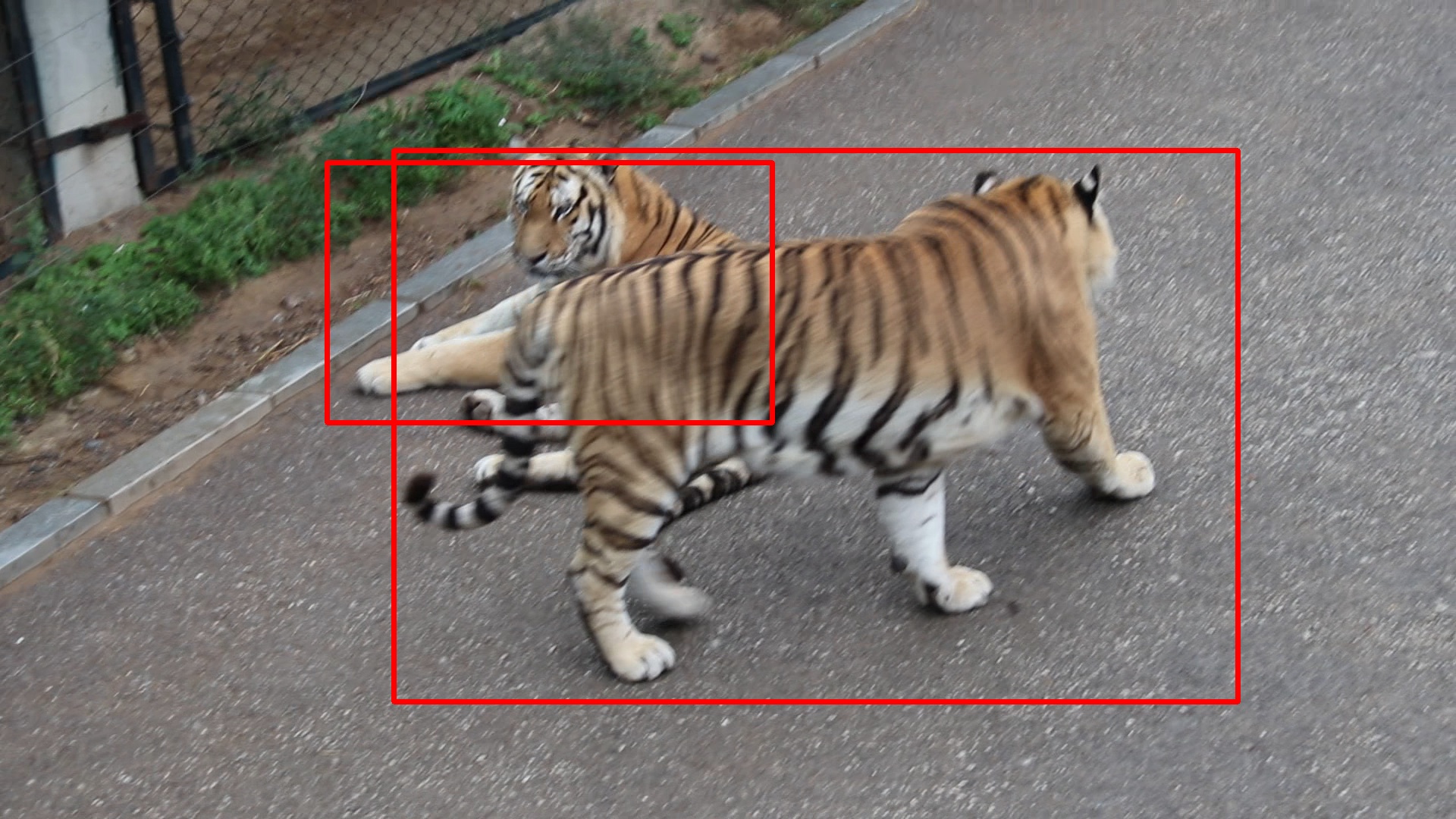} 
    \end{tabular}
    \captionof{figure}{The examples of the predictions on the test set. Our model is able to correctly detect tigers even in complex scenarios and occlusions with a relatively small number of false positives.}
    \label{fig:teaser}
    \vspace{-1.5mm}
\end{center}%
}]

\begin{figure*}[htb]
  \includegraphics[width=0.99\textwidth]{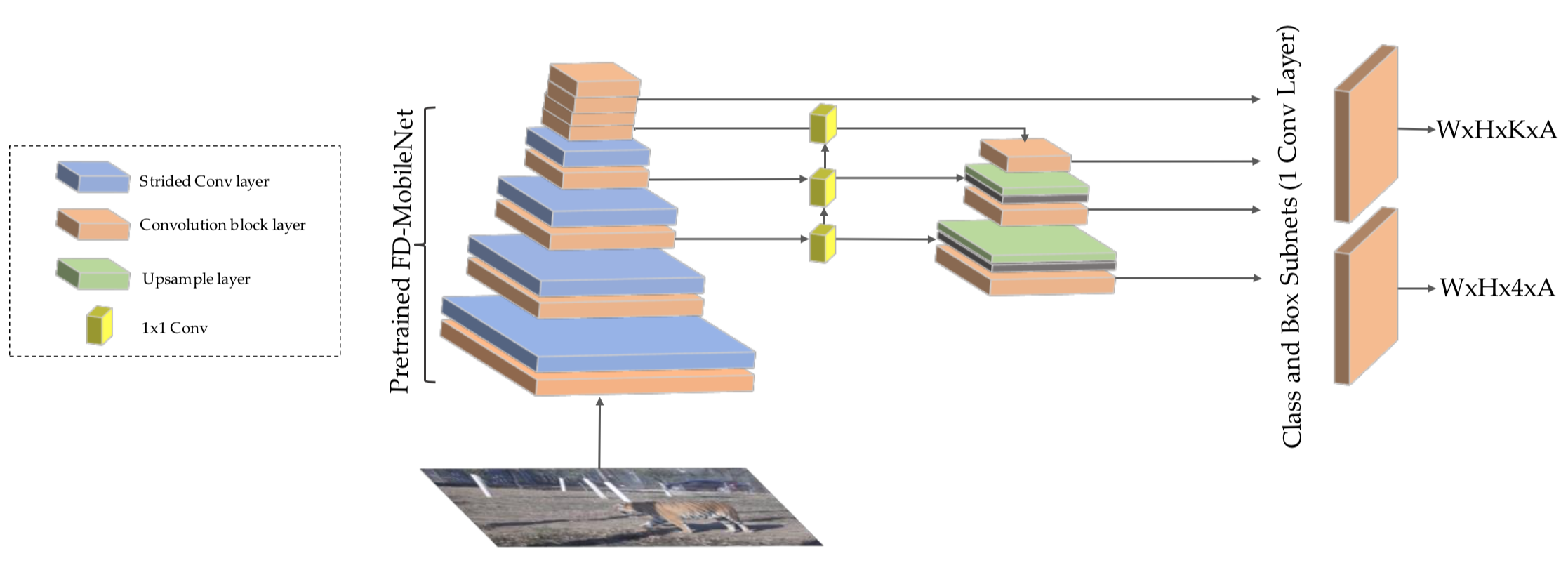}
  \vspace{-1.2em}
  \caption{The proposed architecture with the pretrained FD-MobileNet encoder. We use large feature maps at the last network layers together with the slimmer box regression / classification subnets and Deptwise Separable Convolutions to create an efficient model for edge and mobile devices. In contrast to \cite{retinanet} the Class and Box subnets does not share the weight across the feature maps.}
   \vspace{-1em}
  \label{fig:arch}
\end{figure*}
\begin{abstract}
\vspace{-0.5em}
 The highest \blfootnote{* These two authors contributed equally} accuracy object detectors to date are based either on a two-stage approach such as Fast R-CNN \cite{fastrcnn} or one-stage detectors such as Retina-Net \cite{retinanet} or SSD \cite{ssd} with deep and complex backbones. In this paper we present TigerNet - simple yet efficient FPN based network architecture for Amur Tiger Detection in the wild. The model has \textbf{600k parameters}, requires \textbf{0.071 GFLOPs} per image and can run on the edge devices (smart cameras) in near real time. In addition, we introduce a two-stage semi-supervised learning via pseudo-labelling learning approach to distill the knowledge from the larger networks \cite{pseudolabels}\cite{salt}. For ATRW-ICCV 2019 tiger detection sub-challenge, based on public leaderboard score, our approach shows superior performance in comparison to other methods. The model and the code are available at : \url{https://github.com/KupynOrest/AmurTigerCVWC}

\end{abstract}

\vspace{-1em}
\section{Introduction}
\label{s:intro}
\newcommand{\ra}[1]{\renewcommand{\arraystretch}{#1}}

Wildlife conservation is critical for maintaining species biodiversity. Failure to protect endangered species on Earth may lead to imbalanced ecosystems and affect environmental health. This mission is increasingly depended on accurate monitoring of the geospatial distribution and population health of these endangered species, especially in the face of poaching and loss of habitats. Traditional methods of attaching transmitters to wildlife are prone to sensor failure, difficulties with scaling to large populations, and impossibility to measure how the wildlife interacts with its environment.
Computer vision techniques are a promising approach to wildlife monitoring, especially with the use of unmanned aerial vehicles or camera traps to collect visual data. In particular, detection and re-identification (re-ID) is a core vision method required to obtain accurate population counts and track wildlife trajectory.
However, the deployment of such systems is hampered by several challenges. First, resource constraints on the edge camera require low-power and accurate tiger detection to trigger the image capture and thus avoid that massive irrelevant image capturing consumes space of storage card and battery life. The recent advances of deep learning has led to significant progress in the object detection field. Still, most of the state of the art methods \cite{retinanet}\cite{ssd} use a multi-stage approach or heavy deep neural network architectures that require significant computational resources. 

We introduce a new lightweight object detection architecture that meets all the needed requirements for edge devices deploymenе with superior efficiency in terms of FLOPs and network size while still achieving comparable accuracy to deeper state-of-the-art methods. Our innovations are summarized as below
\begin{itemize}
\vspace{-0.5em}
    \item We present a novel neural network architecture for object detection. The model is based on the Feature Pyramid Network (FPN \cite{fpn}). To further increase efficiency, we extensively use Depthwise Separable Convolutions (originally introduced in \cite{sepconv} and
    used in Inception models \cite{batchnorm}) and lightweight FD-MobileNet backbone \cite{fdmobilenet}. Further, we modify the original RetinaNet architecture to produce larger feature maps at the last layers that allow us to train the network with high efficiency even on small image sizes that significantly reduce the computational requirements. 
    \vspace{-0.5em}
    \item We introduce the two-stage semi-supervised learning approach to distill information from larger deep neural network architectures. We train deep SE-ResNeXt-101 \cite{seresnext} based RetinaNet and use Pseudo-Label \cite{pseudolabels} approach to distill the knowledge from this network using unlabelled data into small FD-MobileNet model. We show that this approach helps to significantly boost performance and efficiently learn the lightweight model even with a small number of annotated samples.
\end{itemize}

\section{Dataset}
The Amur Tiger population currently has less than 600 wild individuals in the world. Capturing enough image data for free-roaming Amur tigers is infeasible as these tigers have an activity range over hundreds of kilometers. Instead, the training data for tiger detection challenge originate from images taken in multiple wild zoos in an unconstrained setting with time-synchronized surveillance cameras and tripod fixed SLR cameras. The dataset includes 4,434 images of high resolution (1920x1080) with 9,496 bounding boxes. To reduce storage, power, and networking consumption the data contains only a subset of frames with at least one Amur Tiger present. The annotations are provided in PASCAL VOC \cite{pascalvoc} format with bounding boxes assigned to each frame. 
\label{s:dataset}

\section{Model}
The goal is to place tight bounding boxes around tigers from images/videos captured by cameras. The solution should be able to run on edge devices so the model should be optimized in terms of FLOPs and size. In this section, we introduce the novel lightweight FPN-like architecture for object detection and a semi-supervised training approach with pseudo-labels.
\subsection{Architecture}
The architecture we use is a modification of RetinaNet \cite{lin2017feature}. It generates multiple feature map layers which that different semantics and contain better quality information. FPN comprises a bottom-up and a top-down pathway. The bottom-up pathway is the usual convolutional network for feature extraction, along which the spatial resolution is downsampled, but more semantic context information is extracted and compressed. Through the top-down pathway, FPNs reconstruct higher spatial resolution from the semantically rich layers. The lateral connections between the bottom-up and top-down pathways supplement high-resolution details and help localize objects. 

This model is agnostic to the feature extractor backbones choice. We use ImageNet-pretrained networks \cite{imagenet} to improve the generalization of the final model by introducing more general features from the backbone activations. To improve the efficiency of the model we use FD-MobileNet \cite{fdmobilenet} as the backbone network which is an efficient and accurate network for very limited computational budgets (10-140 MFLOPs) and outperforms original MobileNet \cite{mobilenet1} both in terms of accuracy and efficiency. To further reduce the complexity, we introduce Deptwise Separable Convolutions \cite{dsc} instead of regular Convolutional Layer in the FPN. In contrast to the original RetinaNet, we created separate box and regression subnets for each feature map activation and limited the number of the layers in the subnet to 1. This allowed us to additionally decrease the number of FLOPs while stile achieving satisfactory results in terms of mAP.

We also discovered that original RetinaNet or SSD implementations fail to converge on smaller image sizes. As shown in table \ref{T:ablation} the RetinaNet-MobileNetv2 baseline achieve the lowest mAP among all the experiments. Due to the Pooling or Stride Convolutional Layer after each feature map, the final activations have too small spatial resolution thus fail to learn good domain-independent features. For default image size = 300 the final activation map output is only 2x2. We removed poolings from the last layers of the model which let us to train and inference on smaller image sizes which in turn significantly improved the efficiency of the model in terms of FLOPs. The output feature maps have spatial resolution \textit{1/8, 1/16, 1/32, 1/32, 1/32, 1/32} of input image size.

\textbf{Loss Functions:} RetinaNet output consists of several feature maps taken from different layers of the FPN network. Category prediction (background or tiger in our case) and bounding boxes coordinates are retrieved from feature maps at the post-processing step. We use separate loss functions for category classification and bounding box coordinates regression. For classification subnet output we use categorical cross-entropy loss:
$$CE=-\sum_{c=1}^{M}{y_{o, c}log(p_{o, c})}$$
where \textit{M} - number of classes in classification task, \textit{y} - binary indicator (0 or 1) if class label \textit{c} is the correct classification for observation \textit{o}, \textit{p} - predicted probability observation \textit{o} is of class \textit{c}. For negative-positive balancing hard-negative mining strategy was used. According to this strategy, for each positive anchor we select $\eta$ negative anchors with the greatest error. Classification loss equals to zero for all the rest negative anchors. For the final model we use $\eta=3$.

For box regression subnet smooth $L_1$ loss \cite{smoothl1} is used:
$$ smooth_{L_1} = \begin{cases} 0.5x^2, & \mbox{if } |x| < 1 \\ |x| - 0.5 & \mbox{otherwise} \end{cases}$$
where \textit{x} - residuals between ground truth and prediction values.

\subsection{Semi-Supervised Learning using Pseudo-Labels}
As in many other domains and applications, the task of this challenge is to create an efficient and lightweight model using a limited number of training samples. Due to the small size of the Amur Tiger population, it is complicated to create a large-scale dataset with various conditions (day time, weather, zones, scales, etc.) Thus, it is crucial to be able to learn from unlabelled data which is usually easier to find and collect/create. We propose to distill the knowledge from the larger networks via pseudo-labelling \cite{pseudolabels} on unlabelled data. This both helps to increase the generalization of the smaller model and utilize the information from the additional data. In particular, we train a large SE-ResNeXt-101 RetinaNet-like model on the labeled training set which produces 0.61 mAP. Further, we use the raw predictions on the unlabelled part of the dataset on this model as the ground truth labels along with the labeled training set and train the final FD-MobileNet object detection model. As shown in Section 6, this semi-supervised learning approach with knowledge distillation allows to significantly improve the generalization of the smaller model.
\label{s:model}

\begin{figure*}[!t]
 \vspace{-2em}
	\centering
    \captionsetup{justification=centering}
    \begin{subfigure}[t]{0.24\textwidth}
        \includegraphics[width=\textwidth]{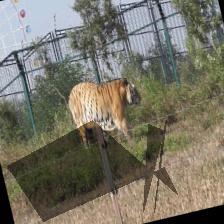}
        \label{fig:11}
    \end{subfigure}
    \begin{subfigure}[t]{0.24\textwidth}
        \includegraphics[width=\textwidth]{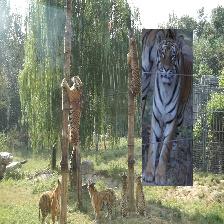}
        \label{fig:12}
    \end{subfigure}
    \begin{subfigure}[t]{0.24\textwidth}
        \includegraphics[width=\textwidth]{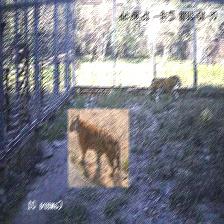}
        \label{fig:13}
    \end{subfigure}
    \begin{subfigure}[t]{0.24\textwidth}
        \includegraphics[width=\textwidth]{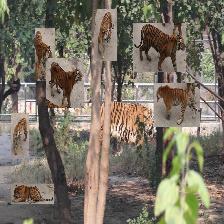}
        \label{fig:14}
    \end{subfigure}
    \vspace{-1em}
    \caption{Network input after augmentations.}
        \vspace{-1.2em}
    \label{fig:aug}
\end{figure*}
\section{Experimental Evaluation}
\subsection{Model Training}
We implemented all of our models using PyTorch~\cite{pytorch}. We take the random 80\% of images as the training set and the left 20\% as the validation set to prevent the model overfitting.
All models were trained on a single 1080-TI GPU, with Adam~\cite{ADAM} optimizer and the learning rate of $10^{-4}$. We use the Early Stopping criterion \cite{earlystop} with mAP on Validation set for 15 epochs. The learning rate is reduced by a factor of 2 each time the validation mAP stopped increasing for 5 epochs. The final model takes 3 hours to converge.

\textbf{Augmentations:} Due to the small size of the training dataset and it's sameness hard augmentations were used in the training process. We used vertical and horizontal flips, affine transforms (rotations, shifts, and scales), Gauss noise, different variants of blur (Gauss blur, median blur, motion blur), random rain and random shadow transforms, colors augmentations (gamma, brightness, contrast). All of the listed transforms were taken from \textit{Albumentations} open-source library \cite{albumentations}.

Additionally we use custom "tiger cutout" augmentation to increase the model performance in cases with multiple tigers present in the frame. We cut out tigers from any images and inserted them on random places of image. \ref{fig:aug} illustrates examples of augmented images.
\textbf{Post Processing:} To improve bounding box coordinates prediction we replaced the standard suppression algorithm with a blending strategy described in Bazarevsky\etal \cite{blazeface}. Also, we implemented test-time augmentations (blending predictions of augmented images) but it gave us poor quality improvement and noticeably increased inference time.


\subsection{Quantitative Evaluation on ATRW dataset}
We compare our model with other participants within the CVWC 2019 challenge on the ATRW dataset \cite{atrw}. Our model is superior to most of the solutions both in terms of Mean Average Precision and FLOPs. On the final mAP per FLOPs metric described above our solution clearly outperforms others. The result is shown in Table\ref{T:atrw}. We achieve high accuracy with the small number of false positives while still maintaining high efficiency. The model can be further optimized using channel pruning and quantization up to 1Mb in size making possible the deployment on any platform or edge device.
\begin{table}[htb]
\small
\vspace{-0.5em}
\caption{The results for CVWC Tiger Detection Challenge. The final metrics are different variants of PPF metric - precision per FLOPs. \url{https://cvwc2019.github.io/leaderboard.html}}
\label{T:atrw}
\centering
\setlength\tabcolsep{3pt}
\begin{tabular}{c|c|c|c}
\toprule
 & mAP & GFLOPs & PPF \\
\hline
\textbf{Our solution} & 0.515 & 0.071 & \textbf{0.630} \\
\hline
dcyhw (FCOS detector) & 0.586  & 5.68 & 0.467 \\
\hline
zdi (Faster RCN + HRN) & 0.601  & 245.3 & 0.388 \\
\hline
lazy-learners (FasterRCNN) & 0.546  & 112.48 & 0.388 \\
\hline
CVWC Team (SSD-v2) & 0.476  & 1.25 &  0.277 \\
\bottomrule
\end{tabular}
\vspace{-0.5em}
\end{table}

\subsection{Ablation Study and Analysis}
\begin{table}[htb]
\small
\vspace{-0.5em}
\caption{Ablation Study on the ATRW dataset.}
\label{T:ablation}
\centering
\setlength\tabcolsep{3pt}
\begin{tabular}{c|c|c}
\toprule
 & mAP & GFLOPs \\
\hline
SSD-MobileNet (solution baseline) & 0.426 & 1.2  \\
\hline
RetinaNet-MobileNetv2 & 0.33  & 0.9 \\
\hline + hard augmentations & 0.42 & 0.9 \\
\hline
+ mean NMS & 0.43  & 0.9 \\
\hline
+ slimmer subnet and remove pooling & 0.511  & 0.64 \\
\hline
+ separable convolutions and FD-MobileNet & 0.489  & 0.071 \\
\hline
+ \textbf{pseudo-labels} & \textbf{0.515}  & \textbf{0.071} \\
\bottomrule
\end{tabular}
\vspace{-0.5em}
\end{table}
We perform an ablation study on the effect of specific components of the pipeline. Starting from the original RetinaNet with MobileNet-v2 \cite{mobilenetv2} backbone, we gradually inject our modifications: adding mean NMS, depthwise separable convolutions, FD-MobileNet backbone, slim subnet, pseudo-labeling, etc. The results are summarized in Table \ref{T:ablation}. We can see that all our proposed components steadily improve either efficiency (FLOPs) or accuracy (mAP). In particular, the mean NMS module and large feature maps contribute most significantly in terms of accuracy, while DSC and FD-MobileNet significantly improve efficiency. 

\section{Conclusion}
This paper introduces a new efficient neural network architecture and training pipeline for object detection. The introduced model can work in real-time and be deployed on edge/mobile devices while still maintaining high accuracy. The proposed solution achieves superior results on the CVWC challenge and had a potential application to wildlife conservation.


{\small
\bibliographystyle{ieee_fullname}
\bibliography{egbib}
}
\end{document}


\title{DeblurGAN-V2: A New Deblurring Framework with Superior Quality,\\ Efficiency, and Flexibility\\ Supplementary Material}

\author{First Author\\
Institution1\\
Institution1 address\\
{\tt\small firstauthor@i1.org}
\and
Second Author\\
Institution2\\
First line of institution2 address\\
{\tt\small secondauthor@i2.org}
}
\maketitle
\section{Real-Time Video Deblurring Result}
To demonstrate the efficiency of out solution, we provide the results of our DeblurGAN-V2 (FPN-MobileNet) model on short video sequences from DVD test dataset. Even though our method is not designed for video deblurring and does not use temporal information, it does not suffer from jittering or any other artifacts is stable and up to 100 times faster than the nearest competitors. The efficiency of out pipeline and MobileNet-V2 backbone allows to achieve up to 30fps on GPU for HD video which opens the possibility of real-time video deblurring and stabilization. For all of the videos in supplementary archive on the left-hand there is an original blurry video and on the right-hand side the stabilized output from our model.

\section{Qualitative Result on GoPro dataset}
On the next page we include more visual results of out DeblurGAN-V2 (FPN-Inception) model on GoPro test dataset. We select the frames from different sequences to demostrate an efficiency of our method in various cases. It shows that DeblurGAN-V2 can handle multiple sources and types of the blur, is stable and artifact-free.
\begin{figure*}[htb]
  \includegraphics[width=0.99\textwidth]{sgopro_1}
  \includegraphics[width=0.99\textwidth]{sgopro_2}
  \includegraphics[width=0.99\textwidth]{sgopro_3}
  \includegraphics[width=0.99\textwidth]{sgopro_4}
  \includegraphics[width=0.99\textwidth]{sgopro_5}
  \label{fig:arch}
\end{figure*}
\begin{figure*}[htb]
  \includegraphics[width=0.99\textwidth]{sgopro_6}
  \includegraphics[width=0.99\textwidth]{sgopro_7}
  \includegraphics[width=0.99\textwidth]{sgopro_8}
  \includegraphics[width=0.99\textwidth]{sgopro_9}
  \includegraphics[width=0.99\textwidth]{sgopro_10}
  \caption{On the left-hand side - blurry image, on the right-hand side - the output of DeblurGAN-V2(FPN-Inception) model.}
  \label{fig:arch}
\end{figure*}